\documentclass[conference]{IEEEtran}
\IEEEoverridecommandlockouts
\usepackage{cite}
\usepackage{subcaption}
\usepackage{amsmath,amssymb,amsfonts}
\usepackage{algorithmicx}
\usepackage{graphicx}
\usepackage{textcomp}
\usepackage{balance}
\usepackage{xcolor}
\usepackage[bookmarks=false]{hyperref}
\def\BibTeX{{\rm B\kern-.05em{\sc i\kern-.025em b}\kern-.08em
    T\kern-.1667em\lower.7ex\hbox{E}\kern-.125emX}}

\usepackage{listings}
\usepackage{multirow} 

\begin{document}
\title{\LARGE Fine-Tuning \textit{Surrogate Gradient Learning} for Optimal Hardware Performance in Spiking Neural Networks}

\author{Ilkin Aliyev and Tosiron Adegbija\\
Department of Electrical and Computer Engineering\\
The University of Arizona, Tucson, AZ, USA\\
Email: \{ilkina, tosiron\}@arizona.edu

}

\maketitle

\begin{abstract}
The highly sparse activations in Spiking Neural Networks (SNNs) can provide tremendous energy efficiency benefits when carefully exploited in hardware. The behavior of sparsity in SNNs is uniquely shaped by the dataset and training hyperparameters. This work reveals novel insights into the impacts of training on hardware performance. Specifically, we explore the trade-offs between model accuracy and hardware efficiency. We focus on three key hyperparameters: surrogate gradient functions, \texttt{beta}, and \texttt{membrane threshold}. Results on an FPGA-based hardware platform show that the \texttt{fast sigmoid} surrogate function yields a lower firing rate with similar accuracy compared to the \texttt{arctangent} surrogate on the SVHN dataset. Furthermore, by cross-sweeping the \texttt{beta} and \texttt{membrane threshold} hyperparameters, we can achieve a $48\%$ reduction in hardware-based inference latency with only $2.88\%$ trade-off in inference accuracy compared to the default setting. Overall, this study highlights the importance of fine-tuning model hyperparameters as crucial for designing efficient SNN hardware accelerators, evidenced by the fine-tuned model achieving a $1.72\times$ improvement in accelerator efficiency (FPS/W) compared to the most recent work.

\end{abstract}

\section{Introduction}
Recent studies have demonstrated significant benefits to considering sparsity in improving hardware efficiency in Spiking Neural Networks (SNNs) \cite{yin2022sata, wang2023spiking}. For example, Yin et al. \cite{yin2022sata} showed that by explicitly exploiting the sparse gradients (as high as $93\%$ for some datasets) in hardware, training SNNs can consume up to $5.58\times$ less energy compared to training on the same hardware without considering sparsity. Similarly, Wang et al. \cite{wang2023spiking} achieved a $2.1\times$ improvement in inference efficiency by exploiting sparsity in hardware compared to sparsity-oblivious hardware. The primary driving factor in the formation of the sparsity characteristic is the input coding scheme of the dataset. Recognizing that this is an active area of research and various encoding approaches are being proposed to reduce the firing rate of the network, in this study, we explore, for the first time, how training hyperparameters can influence the sparsity of SNN models and, in effect, the hardware performance of SNN accelerators.

Surrogate gradients \cite{neftci2019surrogate} are typically employed to train SNN models due to their ability to overcome the non-linear nature of spiking neurons (i.e., binary activations/spikes instead of linear continuous activations). The surrogate gradient approximates the true derivative, thereby effectively providing state-of-the-art accuracy in classification-oriented machine learning (ML) tasks. Nonetheless, the choice of a surrogate function (or the scaling factor of its derivative) is not standardized. This raises the question of how its implementation affects the network's classification accuracy and firing intensity.

To investigate our hypothesis, we systematically study two well-known surrogate functions: \texttt{arctangent} and \texttt{fast sigmoid}, empirically evaluating each function's optimal trade-off points for learning performance versus sparsity under a range of \textit{derivative factors}. We also rigorously evaluate the impact of two critical hyperparameters: \texttt{beta} ($\beta$) and \texttt{threshold} ($\theta$). We leverage our in-house hardware platform to conduct these hardware experiments. This platform allows efficient ``model-to-hardware" mapping by accounting for the model's layer-wise workload characteristics. To the best of our knowledge, this work is the first to present a holistic evaluation of SNN hardware from a training perspective.

\section{Background}
\subsection{Spiking Neuron Model}
The spiking neuron model used is a leaky
integrate-and-fire (LIF) neuron, whose characteristics are shown in Equations \ref{ref:lif1} and \ref{ref:lif2}.

\vspace{-5mm}
\begin{equation} \label{ref:lif1}
    u_{j}[t+1] = \beta u_{j}[t] + \sum_{i} w_{ij} s_{i}[t] - s_{j}[t] \theta
\end{equation}

\begin{equation}\label{ref:lif2}
s_{j}[t] = 
\begin{cases} 
1, & \text{if } u_{j}[t] > \theta \\
0, & \text{otherwise}
\end{cases}
\end{equation}

Here, $\beta$ represents the decay or leak factor of the neuron's membrane potential, typically ranging between 0 and 1. This parameter influences how the previous potential $u_{j}[t]$ affects the current potential $u_{j}[t+1]$. A higher $\beta$ value implies less decay, enabling the neuron to retain more of its previous state, which can increase the likelihood of firing. $\theta$, on the other hand, represents the threshold value that the neuron's membrane potential must surpass to trigger a firing event, as indicated by a spike $s_{j}[t]$. A lower $\theta$ value reduces the potential required for firing, thereby increasing the neuron's firing frequency.

\subsection{Surrogate Approximation Functions}
Surrogate gradients have emerged as a competitive solution to effectively approximate the step function \cite{neftci2019surrogate}. Equations \ref{ref:fs} and \ref{ref:atan} show the approximation formulas used for \texttt{arctangent} and \texttt{fast sigmoid}, respectively.

\begin{minipage}{.54\linewidth}
\vspace{-7mm}
\begin{equation} \label{ref:atan}
    S \approx \frac{1}{\pi} \arctan\left(\pi U \frac{\alpha}{2}\right)
\end{equation}
\vspace{0.1mm} 
\end{minipage}%
\begin{minipage}{.41\linewidth}
\vspace{-7mm}
\begin{equation} \label{ref:fs}
    S \approx \frac{U}{1 + k|U|}
\end{equation}
\vspace{0.2mm}
\end{minipage}

where \( U \) is the membrane potential and \( \alpha \) and \( k \) are derivative scaling factors for each surrogate gradients. %In general, the higher theey are  

\section{Surrogate Gradient Fine-Tuning and Hyperparameter Search Results}
\subsection{Experimental Setup}
We used snnTorch \cite{eshraghian2023training} to construct the spiking neuron models and  PyTorch for training the network using the Street View House Numbers (SVHN) dataset. The network is a convolutional SNN with the following structure: 32C3-P2-32C3-MP2-256-10 (where $X$C$Y$ stands for $X$ filters with size $Y\times Y$ and MP$Z$ for maxpooling with size of $Z \times Z$). We employ cosine annealing \cite{loshchilov2016sgdr} for the learning rate scheduler (with epochs set to 25) during training due to its ability to rapidly converge to optimal accuracy. The trained model was mapped to an in-house hardware platform\footnote{Publicly available at \url{https://github.com/githubofaliyev/SNN-DSE}}, developed in SystemVerilog, and implemented on a Xilinx Kintex® UltraScale+™ FPGA. This hardware efficiently allocates platform resources for the model by leveraging the model’s layer sizes and layer-wise sparsity characteristics to achieve an ultra-low power resource allocation scheme. In addition, the hardware operates in a layer-wise lock-step manner to save memory resources and achieve high throughput. % at the cost of latency.  

\begin{figure}[t!]
    \centering
    \begin{subfigure}{.5\linewidth} % Adjust the width to fit your requirement
        \centering
        \includegraphics[width=\linewidth]{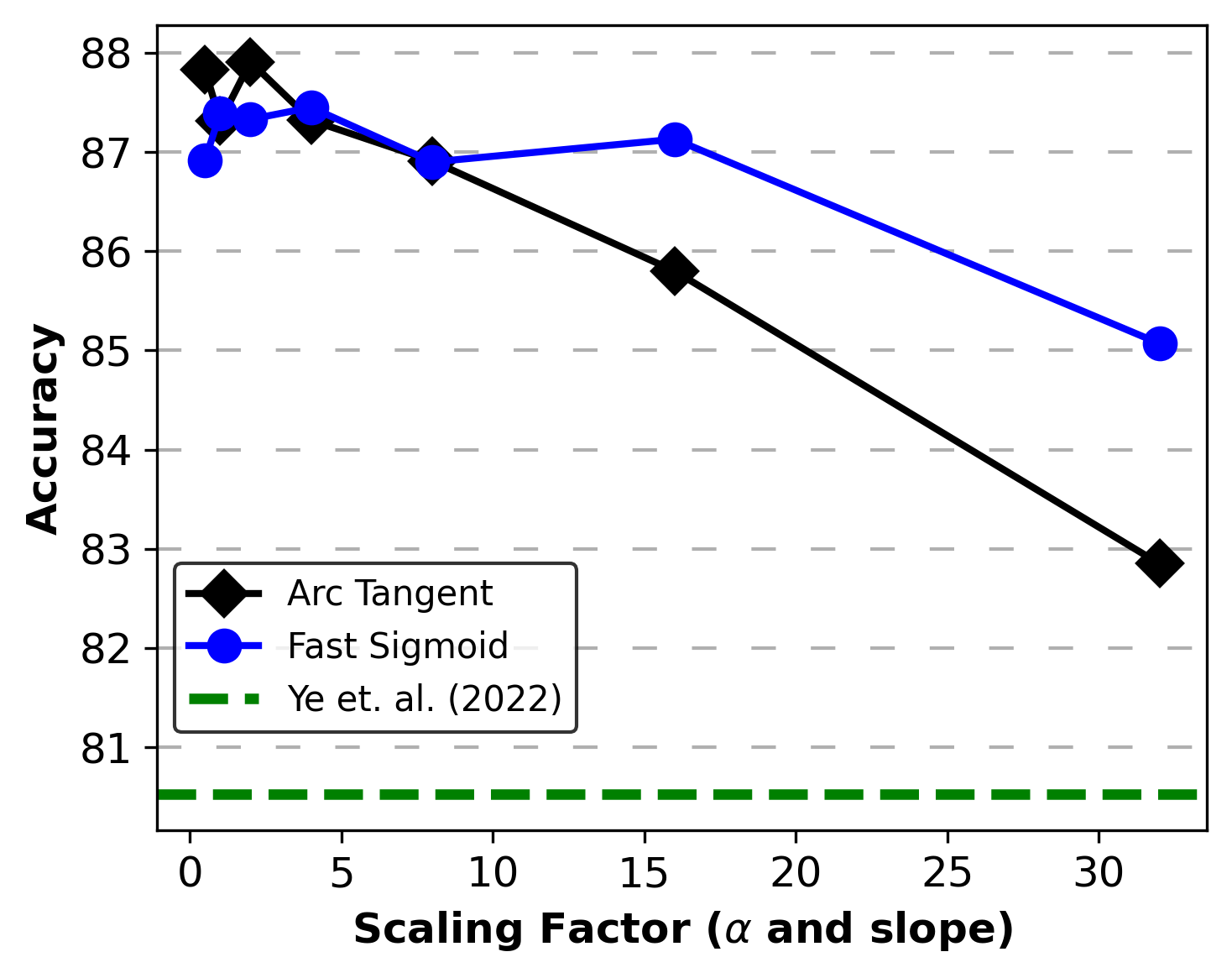}
        %\caption{Caption for Figure A}
        \label{fig:surrogate_acc}
    \end{subfigure}%
    \begin{subfigure}{.5\linewidth} % Adjust the width to fit your requirement
        \centering
        \includegraphics[width=\linewidth]{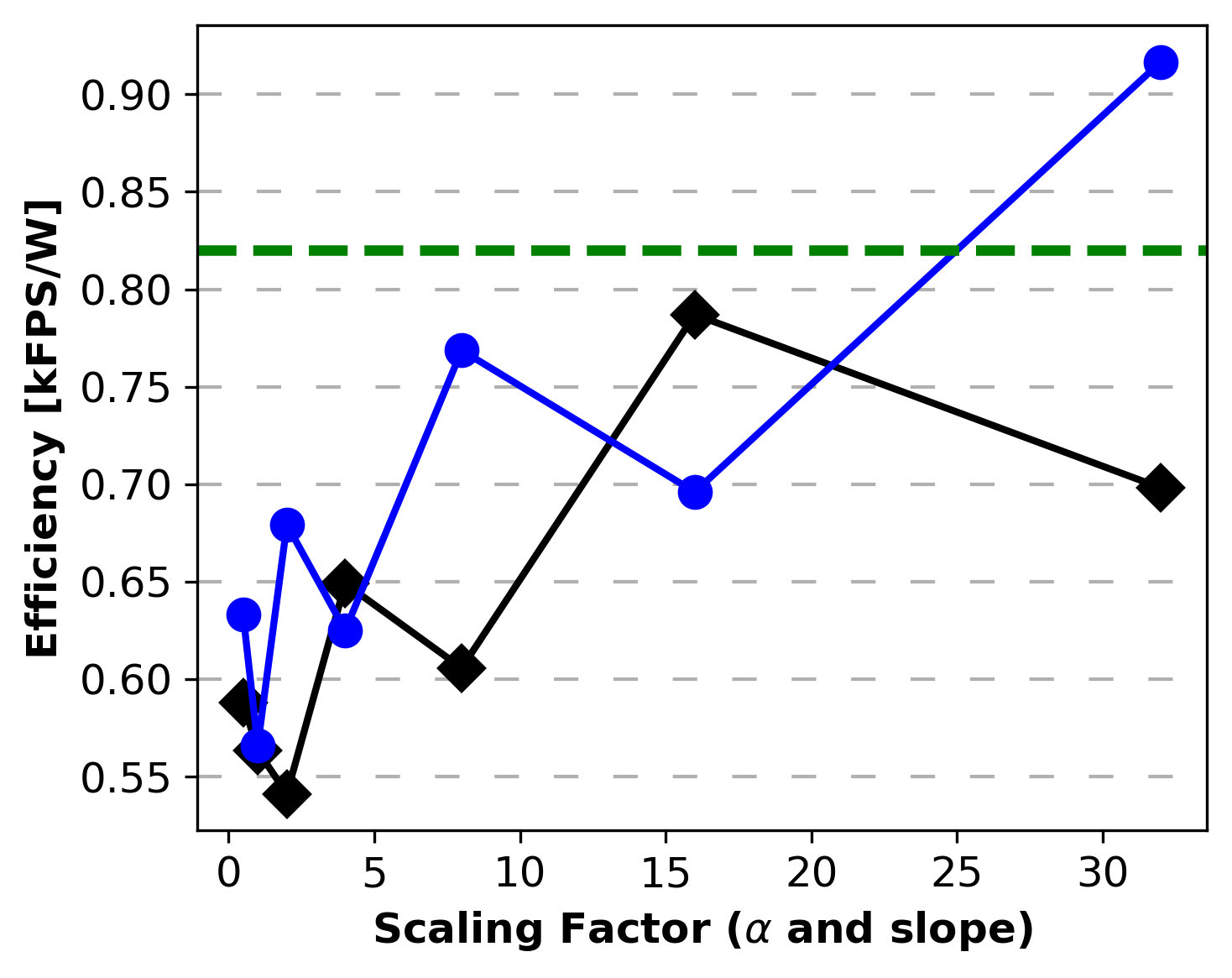} % Replace with your second image
        %\caption{Caption for Figure B}
        \label{fig:surrogate_eff}
    \end{subfigure}
    \vspace{-30pt}
    \caption{Cross-comparison results for \texttt{arctangent} and \texttt{fast sigmoid} surrogate functions over varying derivative scaling factors}
    \label{fig:surrogate}
\end{figure}

\subsection{Results}
\noindent\textbf{Surrogate functions:} To systematically evaluate each surrogate function, we performed a parameter sweep over the derivative scaling factors \( k \) and \( \alpha \) while leaving both $\beta$ and $\theta$ set to their default values ($0.25$ and $1.0$ respectively). Figure \ref{fig:surrogate} shows the variation in accuracy and accelerator efficiency (FPS/W) for each surrogate gradient. We set both \( k \) and \( \alpha \)to the value range of $0.5$ to $32$, beyond which the accuracy for the \texttt{arctangent} surrogate drops below $20\%$. We observe that while both surrogate gradients follow a similar trend in accuracy and efficiency, the \texttt{fast sigmoid} yields lower firing activity (i.e., higher sparsity) compared to the \texttt{arctangent}, resulting in higher accelerator efficiency. Moreover, in both surrogate gradients, through hyperparameter tuning, our network yields higher accuracy than previous work \cite{ye2022implementation} (as highlighted by the horizontal green line in Figure \ref{fig:surrogate}), using the same network architecture and dataset, while the \texttt{fast sigmoid} achieves $11\%$ better accelerator efficiency.

\begin{figure}[t!]
    \centering
    \begin{subfigure}{.5\linewidth} % Adjust the width to fit your requirement
        \centering
        \includegraphics[width=\linewidth]{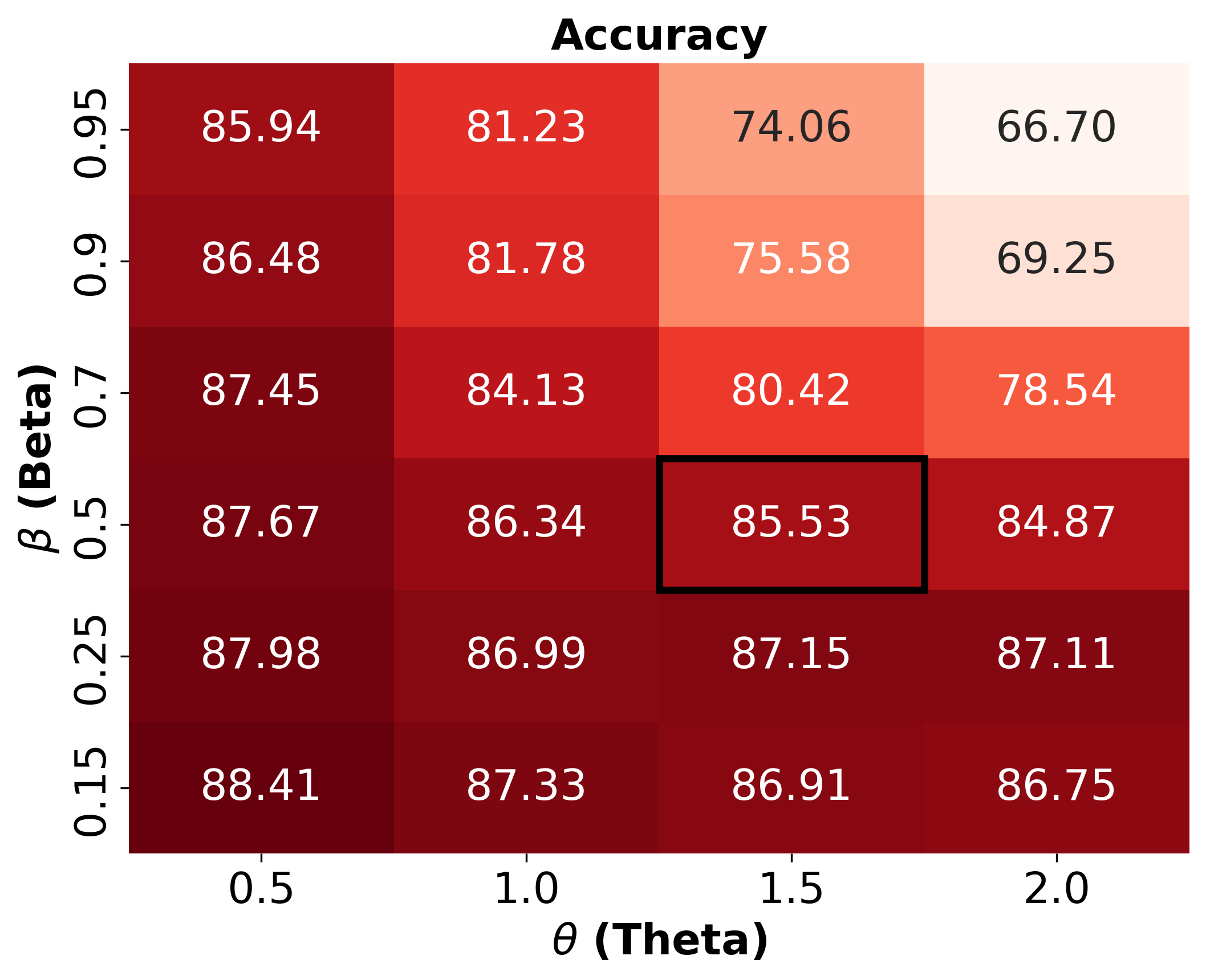}
        %\caption{Caption for Figure A}
        \label{fig:figA}
    \end{subfigure}%
    \begin{subfigure}{.5\linewidth} % Adjust the width to fit your requirement
        \centering
        \includegraphics[width=\linewidth]{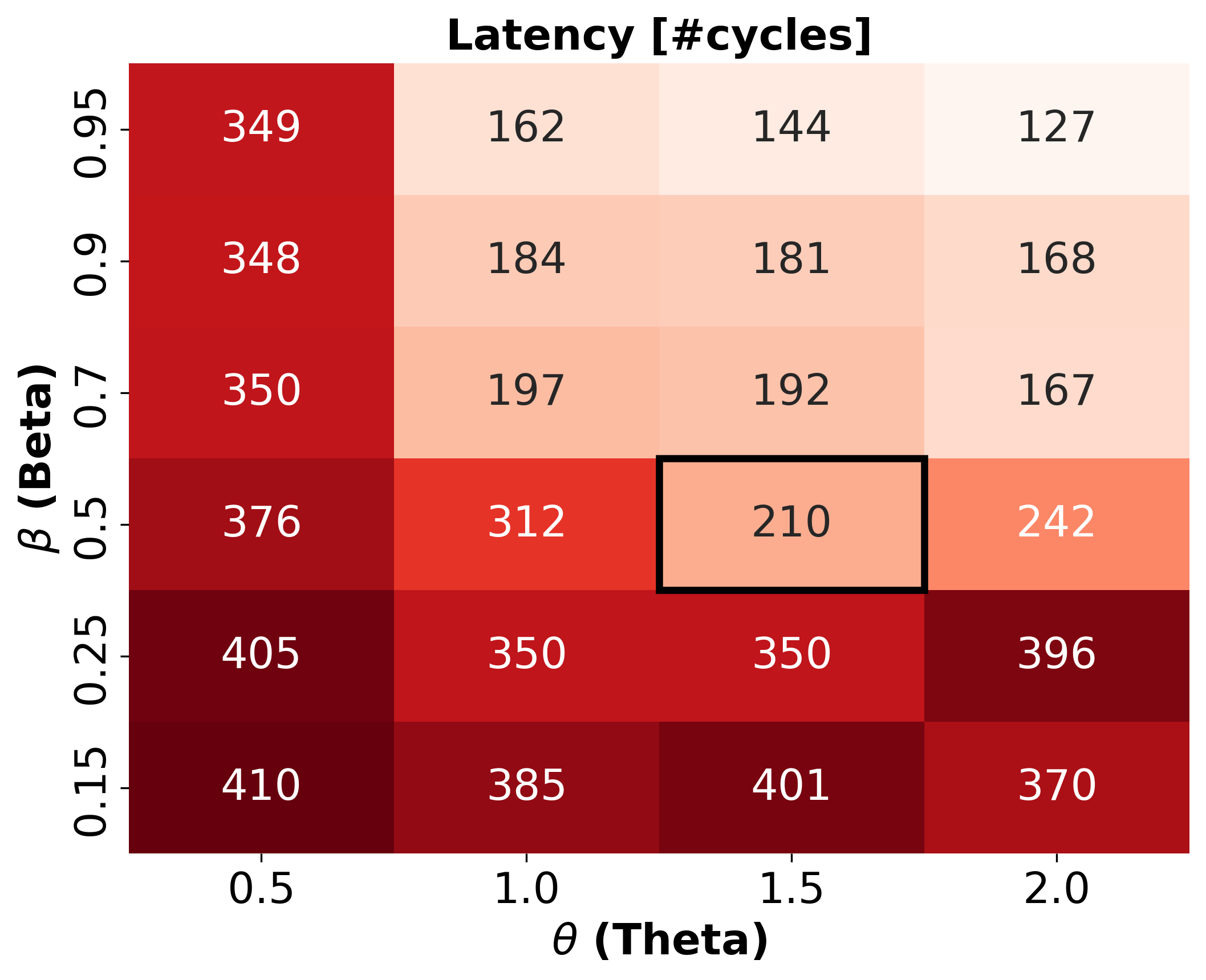} % Replace with your second image
        %\caption{Caption for Figure B}
        \label{fig:figB}
    \end{subfigure}
    \vspace{-30pt}
    \caption{Cross-sweep results for $\beta$ and $\theta$ parameters.}
    \vspace{-10pt}
    \label{fig:beta_thr_sweep}
\end{figure}

\noindent\textbf{Beta-threshold cross sweep:} Given its high sparsity, we chose the \texttt{fast sigmoid} surrogate with a slope scaling factor of $0.25$ for the following experiments. In Figure \ref{fig:beta_thr_sweep}, we cross-sweep $\beta$ (leakage factor) and $\theta$ (threshold potential). Our analysis identifies the optimal balance at a $\beta$ value of $0.5$ and a $\theta$ value of $1.5$. This configuration significantly reduced the inference latency by $48\%$, while only incurring a minor accuracy loss of $2.88\%$, compared to the best accuracy configuration. Compared to prior work \cite{ye2022implementation}, this fine-tuning---with $\beta$ set to $0.7$ and $\theta$ to $1.5$--- achieved $1.72\times$ greater hardware efficiency without degrading the accuracy.

\section{Conclusion}
This study sheds new light on a previously unexplored aspect of SNN hardware accelerator design. While previous research has primarily focused on non-hardware-related factors like dataset encoding, our work pioneers the evaluation of the training hyperparameter space in relation to hardware efficiency. We show that fine-tuning surrogate gradient hyperparameters can provide significant benefits to hardware efficiency and should be considered carefully in the design of efficient SNN accelerators. In future work, we aim to broaden our analysis by exploring additional datasets and the hardware efficiency impacts of other hyperparameters like loss functions. 

\bibliographystyle{ieeetr}
{\small
\bibliography{refs}}
\balance
\end{document}